\documentclass[10pt,twocolumn,letterpaper]{article}

\usepackage[pagenumbers]{cvpr} %

\usepackage[dvipsnames]{xcolor}

\usepackage{graphicx}                      %
\usepackage{tikz}                       %
\usepackage{mathtools} 					%
\usepackage{amssymb}   					%
\usepackage{siunitx}   					%
\usepackage{amsmath}                    %
\usepackage{amsthm}                     %
\usepackage[autostyle]{csquotes}		%
\usepackage{microtype}
\usepackage{algorithm}
\usepackage[noend]{algpseudocode}
\usepackage{comment}

\usepackage[%
  xindy,%
  section = section,%
  nonumberlist,%
  sanitize={symbol=false},%
  shortcuts,%
  acronym,%
  nomain,%
  nowarn%
]{glossaries}

\newacronym{iid}{i.i.d.}{independently and identically distributed}
\newacronym{EBM}{EBM}{Energy-Based Model}
\newacronym{SM}{SM}{Score Model}
\newacronym{NCSM}{NCSM}{Noise Conditioned Score Model}
\newacronym{CNN}{CNN}{convolutional neural network}
\newacronym{EPN}{EPN}{equivariant point network}
\newacronym{VN}{VN}{vector neuron}
\newacronym{VNN}{VNN}{vector neural network}
\newacronym{DL}{DL}{Deep Learning}
\newacronym{DSM}{DSM}{denoising score matching}
\newacronym{SDE}{SDE}{stochastic differentiable equation}

\newacronym{PnP}{PnP}{Perspective-n-Point}
\newacronym{NN}{NN}{neural network}
\newacronym{MLP}{MLP}{multilayer perceptron}

\newacronym{ADD}{ADD}{average distance metric}
\newacronym{AUC}{AUC}{area under the curve}

\newglossary[opg]{operator}{opi}{opo}{List of Operators}
\newglossaryentry{ln}{
  sort        = {natural logarithm},
  name        = {\ensuremath{\mathrm{ln}}},
  symbol      = {\ensuremath{\mathrm{ln}\left(\; \bullet \;\right)} },
  description = {the natural logarithm},
  type        = operator
}

\newglossaryentry{Expmap}{
	sort        = {exponential map},
	name        = {\ensuremath{\mathrm{Expmap}}},
	symbol      = {\ensuremath{\mathrm{ln}\left(\; \bullet \;\right)} },
	description = {exponential map},
	type        = operator
}

\newglossary[syg]{symbol}{sys}{syo}{List of Symbols}

\newcommand{\pdata}[0]{\ensuremath{p_{\mathrm{data}}}}

\newcommand{\SE}[1]{\ensuremath{\mathrm{SE({#1})}}}
\newcommand{\SO}[1]{\ensuremath{\mathrm{SO({#1})}}}

\newcommand{\Expmap}{\ensuremath{\mathrm{Expmap}}}

\newcommand{\rmnum}[1]{\romannumeral #1}

\newcommand{\firstrank}[1]{\textcolor{blue}{#1}}
\newcommand{\secondrank}[1]{\textcolor{orange}{#1}}
\newcommand{\thirdrank}[1]{\textcolor{violet}{#1}}

\newcommand{\unsetfirstuseflag} {
	\glslocalreset{EBM}
	\glslocalreset{NCSM}
	\glslocalreset{VNN}
	\glslocalreset{VN}
	\glslocalreset{SM}
	\glslocalreset{SDE}
	\glslocalreset{DSM}
}

\newcommand\blfootnote[1]{%
  \begingroup
  \renewcommand\thefootnote{}\footnote{#1}%
  \addtocounter{footnote}{-1}%
  \endgroup
}

\definecolor{cvprblue}{rgb}{0.21,0.49,0.74}
\usepackage[pagebackref,breaklinks,colorlinks,citecolor=cvprblue]{hyperref}

\def\paperID{375} %
\def\confName{3DV\xspace}
\def\confYear{2025\xspace}

\title{Particle-based 6D Object Pose Estimation from Point Clouds using Diffusion Models}

\author{Christian M\"oller$^{*}$\\
TU Darmstadt\\
{\tt\small c.moller@outlook.de}
\and
Niklas Funk$^{*}$\\
TU Darmstadt\\
{\tt\small niklas@robot-learning.de}
\and
Jan Peters\\
TU Darmstadt\\
{\tt\small jan@robot-learning.de}
}

\newcommand\scalemath[2]{\scalebox{#1}{\mbox{\ensuremath{\displaystyle #2}}}}
\newcommand{\norm}[1]{\left\lVert#1\right\rVert}
\def\mH{{\boldsymbol{H}}}
\def\mSigma{{\boldsymbol{\Sigma}}}
\def\RR{\mathbb{R}}

\begin{document}
\maketitle
\begin{abstract}
    Object pose estimation from a single view remains a challenging problem.
    In particular, partial observability, occlusions, and object symmetries eventually result in pose ambiguity.
    To account for this multimodality, this work proposes training a diffusion-based generative model for 6D object pose estimation.
    During inference, the trained generative model allows for sampling multiple particles, i.e., pose hypotheses.
    To distill this information into a single pose estimate, we propose two novel and effective pose selection strategies that do not require any additional training or computationally intensive operations.
    Moreover, while many existing methods for pose estimation primarily focus on the image domain and only incorporate depth information for final pose refinement, our model solely operates on point cloud data.
    The model thereby leverages recent advancements in point cloud processing and operates upon an \SE{3}-equivariant latent space that forms the basis for the particle selection strategies and allows for improved inference times. 
    Our thorough experimental results demonstrate the competitive performance of our approach on the Linemod dataset and showcase the effectiveness of our design choices.
Code is available at \url{https://github.com/zitronian/6DPoseDiffusion}.
\end{abstract}

\section{Introduction}
\label{chap:0}

Object pose estimation is a fundamental problem in many applications, including Robotics \cite{ShelvingStackingHanging, DiffusionFields}, Autonomous Driving \cite{KITTI}, and Virtual Reality \cite{augmentedReality}.\blfootnote{$^{*}$ Equal Contribution.}
Despite significant advances in recent years, mainly attributed to learning-based methods, the task remains challenging \cite{hoque2021comprehensive}.
Object pose estimation is particularly challenging in situations in which only a single view of the scene is available.
A single perspective might hide distinct object characteristics and partial observability leads to occlusion that intensifies with the level of clutter \cite{ConfrontingAmbiguity, PoseCNN}.
Additionally, object symmetries further hamper the pose estimation task.
All of the previously described phenomena, i.e., partial observability, occlusion, and symmetric objects eventually result in pose ambiguity and multiple pose hypotheses fitting the observation.
To deal with this ambiguity, this work explores leveraging diffusion models for particle-based object pose estimation.
Generative models \cite{YangSDE, YangGenerativeModelling, TutorialOnEBM, YangHowToTrain} have been shown to excel in learning multi-modal distributions and, therefore, hold promise in addressing the aforementioned challenges.
Additionally, while many existing methods for 6D object pose estimation primarily operate in the image domain \cite{PVNet, hybridpose, PoseCNN} and only incorporate depth information in the final refinement step \cite{deepIm}, in this work, we explore working directly in the 3D point cloud domain.
This work thus aims to align the inherent three-dimensional nature of both the scene and its objects.
Moreover, recent advancements in \gls{NN} architectures specialized for feature extraction from point clouds have opened up exciting possibilities for enhancing pose estimation capabilities from point clouds \cite{VectorNeurons, PointNet, DGCNN}.
In particular, we exploit equivariant feature spaces that naturally emerge from these networks to obtain more expressive encodings and improved inference times.
Lastly, while particle-based approaches to pose estimation naturally yield multiple pose hypotheses, it is also crucial to rank the individual estimates and come up with a final pose estimate.
To this end, we present and compare two novel, simple, but effective particle selection strategies.

This work therefore contributes a novel, particle-based approach for \textbf{(a) 6D Pose Estimation in the Point Cloud Domain.}
Through leveraging information about the 3D model of the object that is to be detected, the inference using a diffusion-based generative model captures the intuition to iteratively move the object model towards the correct position and orientation in the scene.
Moreover, the underlying implicit generative model naturally captures the multimodality that arises from partial observability.
This work also demonstrates effectiveness of \textbf{(b) utilizing \SE{3}-equivariant Vector Neurons.}
We leverage the power of \glspl{VN} to build a \SE{3}-equivariant point cloud encoder.
This creates a latent space that is equivariant to \SE{3} transformations of the input space and yields a meaningful point cloud encoding.
This property enables a substantial inference time improvement with small sacrifices in accuracy.
Lastly, this work introduces \textbf{(c) Novel Pose Selection Methods.}
The particle-based inference process naturally results in multiple pose hypotheses.
We propose two novel strategies to decide upon a single 6D pose from the particles.
One strategy can be deployed for any \gls{SM} whereas the other one leverages the properties of the latent space. Both strategies show high success rates w.r.t. selecting particles resulting in an accurate pose estimation, are computationally efficient, and do not necessitate additional training.
We evaluate our approach throughout extensive experiments conducted on the Linemod dataset \cite{linemodhinterstoisser2011}, demonstrate its competitive performance, and the effectiveness of its individual components.

\section{Related Work}
\label{chap:relatedwork}

The related work section is organized around the topics of point cloud networks, \glspl{SM}, and pose estimation.
They are the key domains addressed and leveraged in this work. 

\label{sec:pointcloudnetworks}

\textbf{Point Cloud Networks.}
Amongst the most popular approaches for \gls{NN}-based point cloud encoding is the PointNet++ \cite{PointNet++} architecture, suitable for learning local structures through hierarchical feature learning.
In another line of work, Hugues et al. \cite{KPConv} define a kernel point convolution for point clouds, which, combined with group convolutions yields \SE{3}-equivariant point networks (EPN) \cite{EPN}.
Yet, EPN's equivariance is only guaranteed up to the degree of its rotation discretization.
Alternatively, Tensor Field Networks \cite{TFN} or \SE{3}-Transformers \cite{TransformerNetworks} provide full \SE{3}-equivariance by design but are cumbersome to integrate and slow \cite{E2PN}.
Deng et al.~\cite{VectorNeurons} present a novel, lightweight \SO{3}-equivariant network based on \glspl{VN}, and demonstrate its straightforward integration into existing architectures, like PoseNet or DGCNN \cite{DGCNN}.
Katzir et al. \cite{ShapePoseDisentanglment} extended the \gls{VN} framework to SE(3)-equivariance.
Following these advances, herein, we also leverage \glspl{VN}~\cite{VectorNeurons} for obtaining expressive point cloud encodings.
Moreover, this work has a particular focus on exploiting the latent's space equivariance for determining the quality of the individual pose hypotheses and accelerating inference, two aspects which remain underexplored in prior work.   

\label{scoremodels}
\textbf{Score Models.}
Yang et al. \cite{YangGenerativeModelling, YangImproved} propose score-based generative modeling, suitable for learning the gradient of the data distribution.
The framework consists of a \gls{NCSM} trained through \gls{DSM} on multiple noise scales.
Once the model is trained, annealed Langevin dynamics is used for generating samples.%
\cite{YangSDE} shows that \glspl{NCSM} can be viewed as an instance of Diffusion Models \cite{DiffusionModels}. %
Following the encouraging results for image generation, recent works trained \gls{NCSM} models for 6D grasp generation in robotics \cite{DiffusionFields} and object rearrangement \cite{ShelvingStackingHanging}.
Inspired by their ability to capture multi-modal distributions, this work introduces a \gls{NCSM} for 6-dimensional object pose estimation in \SE{3}.

\label{poseestimation}
\textbf{6D Pose Estimation.}
In recent years, many deep learning based approaches to 6D pose estimation have been proposed to circumvent manual feature engineering \cite{hoque2021comprehensive}.
A common approach is to train a model extracting image keypoints and subsequently solving PnP for pose estimation.
Exemplarily, the Pixel-wise Voting Network (PVNet)~\cite{PVNet} regresses unit vectors pointing to keypoints, HybridPose \cite{hybridpose} leverages a multi-folded intermediate presentation consisting of keypoints, edge vectors between keypoints, and symmetry correspondences, and YOLO-6D+ \cite{yolo6d+} predicts an object's 3D bounding box and uses the its 2D projections as PnP input.
Alternatively, in PoseCNN \cite{PoseCNN} pose estimation is separated into semantic segmentation, 3D translation estimation, and 3D rotation regression.
Upon availability, the depth image is also exploited for pose refinement using Iterative Closest Point (ICP) \cite{ICP}.
An alternative pose refinement is proposed in Li et al. \cite{deepIm}, where starting from a 3D object model and an initial pose estimate (e.g., from PoseCNN), the pose is refined by rendering the 3D object model into the image, predicting a relative corrective transform, and repeating the process until convergence.
In a similar effort, DenseFusion \cite{DenseFusion} exploits both, the depth and RGB information to obtain pixel-wise features suitable for 6D pose estimation.
In contrast, CloudAAE \cite{gao2021cloudaae} and OVE6D \cite{cai2022ove6d} present approaches solely based on point cloud data.
While the former regresses to the object pose by leveraging an augmented autoencoder, the latter proposes a model for handling arbitrary objects by training on the ShapeNet dataset \cite{chang2015shapenet} and decomposing pose estimation into viewpoint retrieval, in-plane orientation regression and verification.

Closest to this work w.r.t. leveraging generative models for pose estimation are \cite{ConfrontingAmbiguity, xu20246d, jiang2024se, GenPose}.
Hsiao et al. \cite{ConfrontingAmbiguity} use a \gls{NCSM} for 6D pose estimation from images and show that optimizing w.r.t. the \SE{3} group is beneficial compared to treating translation and rotation separately, due to pose ambiguity.
Xu et al. \cite{xu20246d} also present an approach for 6D object pose estimation given RGB image data.
They propose a correspondence-based approach in which the Diffusion Model is tasked with predicting the 2D locations of pre-selected 3D key points. Subsequently, this information is exploited to obtain an object pose estimate by running a PnP solver.
Jiang et al.~\cite{jiang2024se} propose training SE(3)-Diffusion Models for 6D object pose estimation directly from point cloud data.
Compared to this work, they derive a different training objective by following a Bayesian approach, however, they do not consider leveraging equivariant networks, and dealing with multiple pose hypotheses during inference.
The authors of GenPose \cite{GenPose} also build a \gls{NCSM} for category-level 6D object pose estimation from partially observable point clouds.
The problem is approached by training two networks, i.e., an energy model outputting the energy value related to a certain pose and a score model suitable for sampling pose candidates.
The pose candidates holding the highest likelihood are aggregated to obtain a final pose prediction through mean-pooling.
While GenPose \cite{GenPose} deploys a PointNet++ architecture, we make use of \glspl{VN} to encode point clouds and obtain SE(3)-equivariant latent spaces.
Moreover, for final pose estimation, GenPose relies on training an additional network (the energy model).
Our approach instead circumvents the need for extra models by introducing two simple but effective particle selection strategies that do not induce any overhead.

\section{Method}
\label{chap:methods}
\unsetfirstuseflag{}

We proposes learning a point-cloud-based diffusion model for 6D pose estimation.
Herein, a pose $\boldsymbol{H}$ is represented as an element of the group \SE{3}, i.e., $\boldsymbol{H} = \big(\begin{smallmatrix}
  \boldsymbol{R} & \boldsymbol{t}\\
  0 & 1
\end{smallmatrix}\big)$, with rotation $\boldsymbol{R} \in \SO{3}$, and translation $\boldsymbol{t} \in \mathbb{R}^3$. 
After model training, we leverage Langevin dynamics to infer multiple pose candidates.
For each pose candidate, we start with a randomly sampled initial pose which is iteratively refined using our trained model and should eventually end up close to the object's ground truth pose.
To decide upon a final pose, we present two strategies for pose candidate selection, harnessing the properties of our approach, i.e., leveraging the model's score estimate, or alternatively, the point-cloud encoder's latent space.
As observation, we consider a single depth image of the scene, which is converted into a point cloud using the camera's intrinsic.
Additionally, we assume to have access to the 3D model of the object. This allows us to sample points and their corresponding normal vectors.

\subsection{Model Architecture}
\label{sec:model_architecture}
\begin{figure}	
	\centering
	\includegraphics[width=\linewidth]{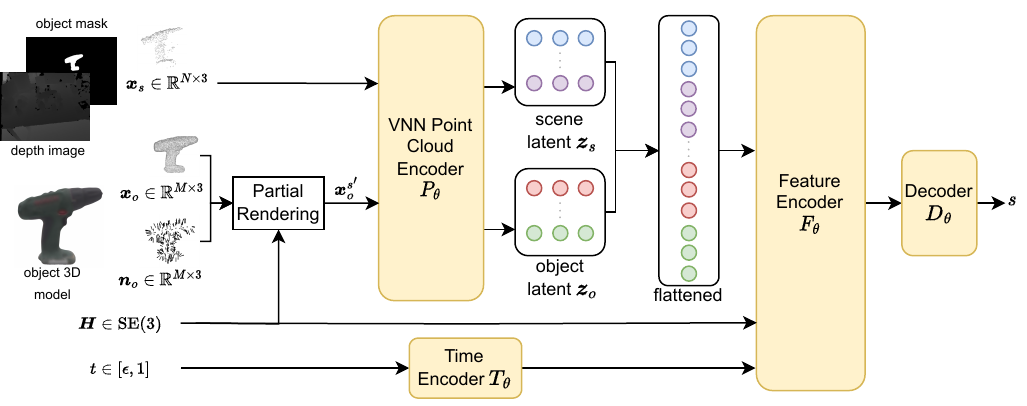}
	\caption{\Gls{NCSM} architecture for learning 6D pose distributions. The object point cloud  $\boldsymbol{x}_o$ is sampled from a 3D model of the object whose pose we want to determine in the scene. Given a specific pose $\boldsymbol{H} \in \SE{3}$ and its normal vectors $\boldsymbol{n}_o$ the object point cloud is partially rendered. Scene point cloud $\boldsymbol{x}_s$ and partial rendered object point cloud $\boldsymbol{x}_{o}^{s'}$ are embedded through a shared encoder that leverages \glspl{VN}. Finally, the scene latent $\boldsymbol{z}_s$ and object latent $\boldsymbol{z}_o$ are flattened and together with an encoded time step and the respective pose $\boldsymbol{H}$ fed through a Feature Encoder $F_{\theta}$ and Decoder $D_{\theta}$ to predict a score $\boldsymbol{s} \in \mathbb{R}^{6}$.}
	\label{fig:methodsModelArchitecture}
\end{figure}

Our proposed model architecture is shown in \cref{fig:methodsModelArchitecture}.
The network receives four inputs:
(\rmnum{1}) The depth image of a partial scene view together with a segmentation mask which is converted to point-cloud form $\boldsymbol{x}_s\in \mathbb{R}^{N \times 3}$.
(\rmnum{2}) A current pose hypothesis $\boldsymbol{H} \in \SE{3}$, which, is to be refined based on the model's output. 
(\rmnum{3}) The time step $t\in [0,1]$ that governs the diffusion process. The time step essentially indicates how advanced the iterative pose refinement process is. More details on the diffusion process follow in \cref{sec:traininginference}.
(\rmnum{4}) A set of points sampled from the fully observable 3D object model $\boldsymbol{x}_o\in \mathbb{R}^{M \times 3}$ along with their normal vectors $\boldsymbol{n}_o\in \mathbb{R}^{M \times 3}$.
Importantly, the object model's point cloud is pre-processed by a partial rendering module to account for the fact that the scene is only partially observable.
Both point clouds, i.e., the scene point cloud and the object point cloud (after passing the partial rendering module) are processed individually by a shared \gls{VNN} encoder to obtain the respective \SE{3}-equivariant latent representations $\boldsymbol{z}_s \in \mathbb{R}^{D \times 3}$ and $\boldsymbol{z}_o \in \mathbb{R}^{D \times 3}$ with dimensions $D$. 
Subsequently, the latents are flattened and concatenated with the rigid transformation $\boldsymbol{H}$, and the time encoding \cite{DiffusionFields} to extract features via the Feature Encoder $F_{\theta}$ represented as a standard \gls{MLP} $F_{\theta}$.
A smaller \gls{MLP} forms the Decoder $D_{\theta}$ and predicts the respective score $\boldsymbol{s} \in \mathbb{R}^6$ for updating the current pose estimate.

Since the partial rendering module and the \gls{VNN} encoder are crucial elements of our approach, we provide a more detailed description in the following.

\textbf{{Partial Rendering Module}.}
\label{subsec:partialrendering}
Based on the current pose hypothesis $\boldsymbol{H}$, the object points are first transformed into the scene via $\boldsymbol{x}_{o}^{s} = \boldsymbol{H}\boldsymbol{x}_o$.
However, without further modifications, the scene view yields a partial point cloud while the object point cloud represents the fully observable object. 
Therefore, $\boldsymbol{x}_{o}^{s}$ will contain points that are located at areas of the surface of the object in the scene that are invisible from the camera's perspective, complicating the pose refinement task.
The partial rendering module aims to account for this mismatch by excluding those points from $\boldsymbol{x}_{o}^{s}$ and creates a partial observable point cloud $\boldsymbol{x}_{o}^{s'}$.
This is achieved by leveraging the normal vector information $\boldsymbol{n}_{o}^{s} = \boldsymbol{H}\boldsymbol{n}_o$.
Our criteria for a point to be visible, i.e., front-facing, is that the normalized dot-product of the vector originating from the camera's viewpoint towards the point $\boldsymbol{p}$ and the points associated surface normal $\boldsymbol{n}$ vector is smaller than some predefined threshold $\epsilon$, i.e., $(\boldsymbol{p} \cdot \boldsymbol{n})/\lVert \boldsymbol{p} \rVert_2 \leq -\epsilon$.
From a geometrical standpoint, we examine the component of the normal vector $\boldsymbol{n}$ w.r.t. the vector $\boldsymbol{p}$ that goes from the camera's origin towards the point and require it to be negative, and smaller than threshold $\epsilon$ in order for the point to be front facing. 
As the introduced process is only a heuristic for the visibility of points from a certain perspective, we want to enforce some robustness against incorrect categorizations of points into visible and not visible. For this reason, the margin $\epsilon$ is sampled from a Gaussian $\epsilon \sim \mathcal{N}(0,0.1)$ for every forward pass and clipped to $[-0.3,0.3]$.
Lastly, we want to remark that this computationally efficient procedure does not account for front-facing points that are occluded by other objects.

\textbf{Point Cloud Encoder.}
\label{subsec:pointcloudencoder}
The point cloud encoder leverages \SO{3}-equivariant \glspl{VN} from \cite{VectorNeurons}, and therefore maps from the point cloud representation to a $D \times 3$ dimensional space, i.e., $P_{\theta} \colon \mathbb{R}^{N \times 3} \to \mathbb{R}^{D \times 3}$ with latent dimension $D$ (\cref{fig:methodsModelArchitecture}).
While this architecture guarantees \SO{3}-equivariance \cite{VectorNeurons}, it lacks equivariance w.r.t. arbitrary translations.
However, as this is desired for our approach, we initially center the input point cloud $\boldsymbol{x} \in \mathbb{R}^{N \times 3}$ using its mean $\bar{\boldsymbol{x}}$, and account for the centering by adding the substracted mean back onto the three dimensional latent points $z \in \mathbb{R}^{D \times 3}$, i.e., $\boldsymbol{z} = P_{\theta}(\boldsymbol{x}-\bar{\boldsymbol{x}}) = P_{\theta}(\boldsymbol{x}-\bar{\boldsymbol{x}}) + \bar{\boldsymbol{x}}$.
This formulation results in a \SE{3}-equivariant latent space, that comes at the cost of only being able to encode rotational information during the point cloud encoding. Since the translational information is missing in the first encoding, it needs to be extracted in the subsequent Feature Encoder $F_{\theta}$.
For the additional details, please see Appendix \cref{app:add_information_point_cloud_enc}.

\subsection{Training and Inference}
\label{sec:traininginference}
We employ a diffusion-based approach for pose estimation and leverage a \gls{SDE} to govern the diffusion process.
Through exploiting the proposed \SE{3}-equivariant latent space, we additionally present a more time-efficient latent-space inference process.

\textbf{Diffusion.}
The object poses $\boldsymbol{H}$ are diffused along infinite noise scales \cite{YangSDE}. In our case, the \gls{SDE} $dx {=} \sigma^{t} dw, t {\in} [0,1]$ governs the diffusion process,
with $w$ being the standard Wiener process \cite{YangSDE}.
Therefore, the perturbed data distribution equates to $q_{t}(\boldsymbol{\hat{H}}) = \int_{\boldsymbol{H}} q_{t}(\boldsymbol{\hat{H}}|\boldsymbol{H})p_{data}(\boldsymbol{H}) d \boldsymbol{H}$ with $q_{t}(\boldsymbol{\hat{H}}|\boldsymbol{H}) = \mathcal{N}(\boldsymbol{\hat{H}}| \boldsymbol{H}, \textrm{var}(\sigma, t) \boldsymbol{I})$, and the variance being $\textrm{var}(\sigma, t) = \frac{1}{2\log(\sigma)} (\sigma^{2t}-1)$.
This is analogous to perturbing the data with $\mathcal{N}(0,\sigma_{1}^{2})$, $\mathcal{N}(0,\sigma_{2}^{2})$, $\cdots$, if $\sigma_1<\sigma_2<\cdots$ is a geometric progression.
In our work we set $\sigma{=}0.5$.
Practically, we sample from the perturbed distribution $q_{t}(\boldsymbol{\hat{H}})$ through first sampling a pose from the data distribution $\boldsymbol{H} \sim \pdata (\boldsymbol{H})$ and composing it with a white noise vector $\boldsymbol{\epsilon} \in \mathbb{R}^{6}$ sampled from the respective noise scale distribution $\boldsymbol{\epsilon} \sim \mathcal{N}(0,\textrm{var}(\sigma, t)\boldsymbol{I}_6)$ \cite{DiffusionFields}. Thus, the noise perturbed pose equates to $\boldsymbol{\hat{H}} = \boldsymbol{H} \Expmap (\boldsymbol{\epsilon})$.
We therefore follow~\cite{sola2018micro, DiffusionFields} in that we work within the vector space $\scalemath{0.9}{\RR^6}$ instead of the Lie algebra.
For moving the elements between Lie Group and vector space, we rely on the logarithmic and exponential maps, i.e., $\scalemath{0.9}{\textrm{Logmap}: \textrm{SE(3)} \xrightarrow{} \RR^6$ and $\textrm{Expmap}: \RR^6 \xrightarrow{} \textrm{SE(3)}}$ respectively~\cite{sola2018micro}.

\textbf{Training.}
In line with \cite{DiffusionFields}, our proposed model (\cref{sec:model_architecture}) is trained using \gls{DSM} with loss term 
\begin{equation}
\label{eq:myloss}
\scalemath{0.85}{
\mathcal{L} = \mathbb{E}_{t \in [\epsilon,1]} \mathbb{E}_{q_{t}(\boldsymbol{H},{\boldsymbol{\hat{H}}})} \left[ \left\Vert s_{\theta}(\boldsymbol{\hat{H}}, t) - \nabla_{\boldsymbol{\hat{H}}} \mathrm{log}(q_{t} (\boldsymbol{\hat{H}}| \boldsymbol{H})) \right\Vert_2^2 \right],
}
\end{equation}
with $\epsilon$ being the smallest time step ($\epsilon=1e^{-5}$ in our case). We follow \cite{YangImproved} and rescale the output of the \gls{NCSM} by $1/\sqrt{\textrm{var}(\sigma, t)}$. 

\textbf{Inference.}
\label{subsec:myinference}
For inferring an object pose estimate, we deploy an inverse Langevin dynamics process, starting from the largest time step $t{=}1$ and a random initial pose $\boldsymbol{H}_{0}=\Expmap(\boldsymbol{\epsilon})$ sampled from $\boldsymbol{\epsilon} \sim \mathcal{N}(0,\textrm{var}(\sigma, 1)\boldsymbol{I}_6)$.
Then follows a sequence of $L$ equally spaced time steps $(t_{i})_{i=0}^{L-1},  t_{i}=(1-(i/(L-1)) (1-\epsilon))$ starting at $t_{0}{=}1$ and ending at $t_{L-1}{=}\epsilon$.
At each time step we update the pose following the Langevin dynamics process $\boldsymbol{H}_{l+1}=  \Expmap(\alpha_{l} s_{\theta}(\boldsymbol{H}_{l}, t_l)+ \sqrt{2\alpha_{l}}0.01\boldsymbol{\epsilon})\boldsymbol{H}_{l}$,
with $\boldsymbol{\epsilon} \sim \mathcal{N}(\boldsymbol{0},\boldsymbol{I}_6)$, the score estimate by our \gls{NCSM} $s_{\theta}(\boldsymbol{H}_{l}, t_l)$ and dynamic step size $\alpha_{l}$.
Note that the pose update is performed in the Lie group through composition \cite{MicroLieTheory}. To enhance exploration in the search space a noise term is added to the predicted score in the Langevin dynamics process.
In the final five iterations of the inference process, this term is abolished leaving $\boldsymbol{H}_{l+1}=\Expmap(\alpha_{l} s_{\theta}(\boldsymbol{H}_{l}, t_l))\boldsymbol{H}_{l}$. 

\textbf{Observation-encoding during Inference.}
As shown in Figure \ref{fig:methodsModelArchitecture}, our proposed architecture encodes the scene and object point cloud individually.
Since the scene point cloud $\boldsymbol{x}_s$ does not change during inference, we only need to encode it once, i.e., $\boldsymbol{z}_s = P_{\theta}(\boldsymbol{x}_s)$.
However, since the object pose estimate is updated iteratively, and since we are dealing with partial observability, the partial rendering module results in a changing object point cloud $\boldsymbol{x}_o$ in every iteration.
Instead of rendering and re-encoding in every iteration, to improve the inference time, we can utilize the \SE{3}-equivariant latent space, and simply transform the object latent $\boldsymbol{z}_{o}$, at the cost of still relying upon the prior partial rendering.
Thus, during inference, we add anhother hyperparameter $k$, which determines that the object point cloud is encoded and re-rendered only in every $k^{\mathrm{th}}$ iteration, and in between, we solely rotate the object latent for improved efficiency.
This results in a trade-off between inference speed and accuracy.
If inference time is not the bottleneck of the application, one could choose $k=1$, which corresponds to re-rendering and encoding the object point cloud in every iteration.
We provide the pseudocode in Appendix \cref{app:inference_pseudocode}

\subsection{Pose Estimation through Particle Selection}
\label{sec:twostrategies} %
The inference process as described in \cref{subsec:myinference} can be used to generate multiple samples, leading to a set of $N$ pose hypotheses (particles) $\mathcal{H}=\{ \boldsymbol{H}_1, \boldsymbol{H}_2, \cdots, \boldsymbol{H}_N\}$.
To condense these hypotheses to one pose prediction $\boldsymbol{H}^*$, we introduce two techniques.

\paragraph{Selection By Score.} 
\label{sec:selectionbyscore}
The inference process as presented in \cref{subsec:myinference} generates a history of $L$ score values $\{{\boldsymbol{s}_0, \boldsymbol{s}_1}, \cdots, \boldsymbol{s}_{L-1}\} \in \mathbb{R}^{L \times 6}$.
As shown in \cite{YangGenerativeModelling}, for score matching in Euclidean Spaces, i.e., with $q_{t} (\boldsymbol{\hat{x}}| \boldsymbol{x})=\mathcal{N}(\boldsymbol{\hat{x}}| \boldsymbol{x}, \sigma_t)$, the objective for score matching equates to $\nabla_{\boldsymbol{\hat{x}}} \mathrm{log}(q_{t} (\boldsymbol{\hat{x}}| \boldsymbol{x})) = - (\boldsymbol{\hat{x}}-\boldsymbol{x})^2 / \sigma_t^2$.
For our case of score matching in \SE{3}, the Gaussian is defined according to $\scalemath{0.9}{q_t(\hat{\mH} | \mH, \mSigma) \propto \exp \left(-0.5~ \norm{\textrm{Logmap}(\mH_{\mu}^{-1} \mH)}_{\mSigma^{-1}}^2  \right)}$, centered around its mean $\scalemath{0.9}{\mH\in SE(3)}$ and with covariance matrix $\scalemath{0.9}{\mSigma \in \RR^{6\times 6}}$~\cite{chirikjian2014gaussian}.
Therefore, for score matching in \SE{3}, the score should match $\nabla_{\boldsymbol{\hat{H}}} \mathrm{log}(q_{t} (\boldsymbol{\hat{H}}| \boldsymbol{H})) \propto \textrm{Logmap}(\mH^{-1} \hat{\mH})$, which essentially is a distance vector within the vector space $\RR^{6}$.
From these derivations, it is clear that the score, and in particular its 2-norm, i.e., $\left\Vert \boldsymbol{s} \right\Vert_2$ is an indicator of how close the current sample is to a sample from the dataset.
Therefore, as a first heuristic to rank the particle-based pose hypotheses, we will consider the last score value, as for precise pose estimates it should be smaller.

\paragraph{Selection By Latent.} 
\label{sec:selectionbylatent}
The partially rendered object point cloud and the cropped scene point cloud are encoded using the same point cloud encoder $P_{\theta}$.
Therefore, the selection by latent follows the geometric intuition, that the quality of pose predictions correlates with the proximity of the object latent $\boldsymbol{z}_o \in \mathbb{R}^{D \times 3}$ to the scene latent $\boldsymbol{z}_s \in \mathbb{R}^{D \times 3}$ given a specific pose $\boldsymbol{H} \in \SE{3}$.
We want to underline that the output of the encoder are $D$ 3-dimensional points, which we will compare in terms of euclidean distance.
In particular, the scene latent consists of the vectors $\boldsymbol{z}_s= \{ {}^{(0)}\boldsymbol{z}_s, {}^{(1)}\boldsymbol{z}_s, \cdots,{}^{(D-1)}\boldsymbol{z}_s \}$.
The object latent on the other hand varies along the $N$ different particles.
To formalize this selection strategy we define a proximity function $\mathrm{lprox} \colon \mathbb{R}^{D \times 3} \to \mathbb{R},\quad \mathrm{lprox}(\boldsymbol{z}_{o}, \boldsymbol{z}_{s}) = \sum_{i=0}^{D-1} \left\Vert {}^{(i)}\boldsymbol{z}_s - {}^{(i)}\boldsymbol{z}_o \right\Vert_2$ that can be used to determine the proximity of the $N$ pose hypotheses, i.e., object latents w.r.t. the scene latent.
Under this prediction strategy, the final predicted pose is the particle that minimizes this distance.

\newcommand{\AUC}{AUC}
\newcommand{\ADDACC}{ACC-0.1}

\section{Evaluation Procedure}
\label{chap:experiments}
\unsetfirstuseflag{}
This section provides all the details regarding the evaluation of our proposed approach for 6D pose estimation.

\textbf{Dataset.}
\label{sec:dataset}
\begin{figure}	
	\centering
	\includegraphics[width=0.8\linewidth]{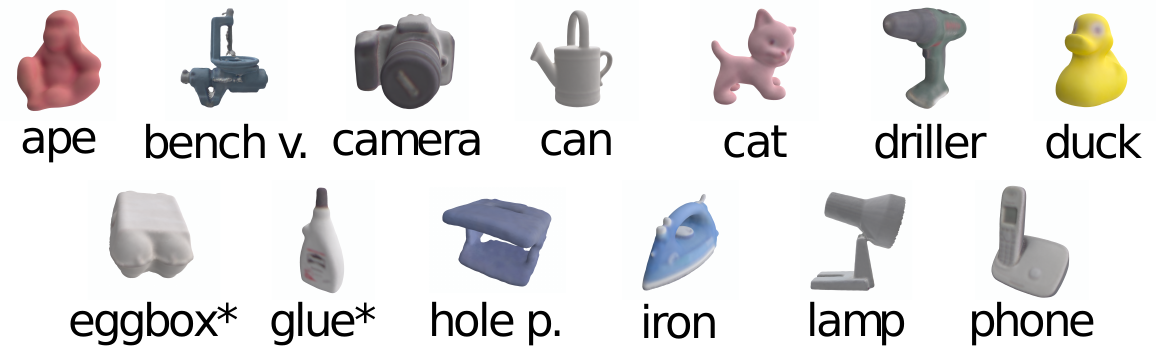}
	\caption{Objects of the Linemod dataset \cite{linemodhinterstoisser2011} excluding the two objects with incomplete 3D models. * denotes symmetric objects.}
	\label{fig:experimentsLMobjects}
\end{figure}
We evaluate our method on the  Linemod dataset \cite{linemodhinterstoisser2011}, which consists of 15 texture-less objects and around 1200 images per object.
In line with many other works on 6D pose estimation such as \cite{UncertaintyDriven, efficientpose, yolo6d+, hybridpose, DenseFusion}, we only evaluate on the 13 objects for which complete 3D models are provided (cf. \cref{fig:experimentsLMobjects}), and adopt their train test split in which 200 images per object are used for training, and 1000 for testing.
If not stated differently, all results are averaged over all test samples.

\textbf{Metrics.}
\label{sec:metrics}
We use the \gls{ADD} \cite{HinterModelBasedTraining} and the ADD-S metric for evaluating the quality of the pose estimates for non-symmetric and symmetric objects, respectively.
Going forward we will only refer to the \gls{ADD} which implies the computation of the ADD-S in cases of symmetric objects. We follow the convention to classify a pose prediction as correct if the \gls{ADD} is less than 10\% of the objects diameter which gives the accuracy metric \ADDACC{}, which we solely denote as Accuracy. Varying this threshold results in a accuracy-threshold curve, which allows us to compute the \gls{AUC} \cite{PoseCNN}.

\textbf{Model Implementation Details.}
\label{sec:implementationDetails}
We equip the model with a latent space of size 282, i.e., $D{=}94$ for both, encoding the object and the scene.
The Feature Encoder $F_{\theta}$ consists of seven fully connected layers of size 512 with 20\% dropout per layer.
The Decoder $D_{\theta}$ processes the concatenated latents with three fully connected layers (size 256) and outputs the score value $\boldsymbol{s} \in \mathbb{R}^6$.
To obtain the object point cloud, we sample 1024 points from the set of face centroids of the 3D model and the face's normals as the respective normal vectors.
The scene point cloud is cropped based on the segmentation mask of the visible part of the object, provided by the dataset.
We also sample 1024 points from the scene point cloud.
The point clouds are originally in mm, we scale them by 500 such that one unit equals 50cm.

\textbf{Model Training.}
We train a single model for all 13 objects.
The model is trained for 3500 epochs with AdamW optimizer and a cosine annealing scheduler with a maximal learning rate of 0.001 and a maximum number of 500 iterations.
We choose a batch size of 4 whereas each sample is pertubed with 4 noise scales leading to an effective batch size of 16.
Training of the model on the entire Linemod dataset takes $\approx$80 hours on a NVIDIA V100 GPU with 32GB. 
Lastly, we want to highlight that we only train on the 200 training images per object from the dataset split and do not add additional synthetic training data.

\textbf{Model Inference.}
If not stated otherwise, the standard inference configuration performs 100 iterations of Langevin dynamics using 20 randomly sampled particles.
To achieve the best accuracy the object point cloud is rendered every iteration, setting $k{=}1$ (see also \cref{alg:ourannealed} in the Appendix).

\section{Experiments}
\label{sec:experiments}

\subsection{Selection Methods}
\label{sec:selection_methods}

\begin{figure}
	\centering
		\includegraphics[width=.75\linewidth]{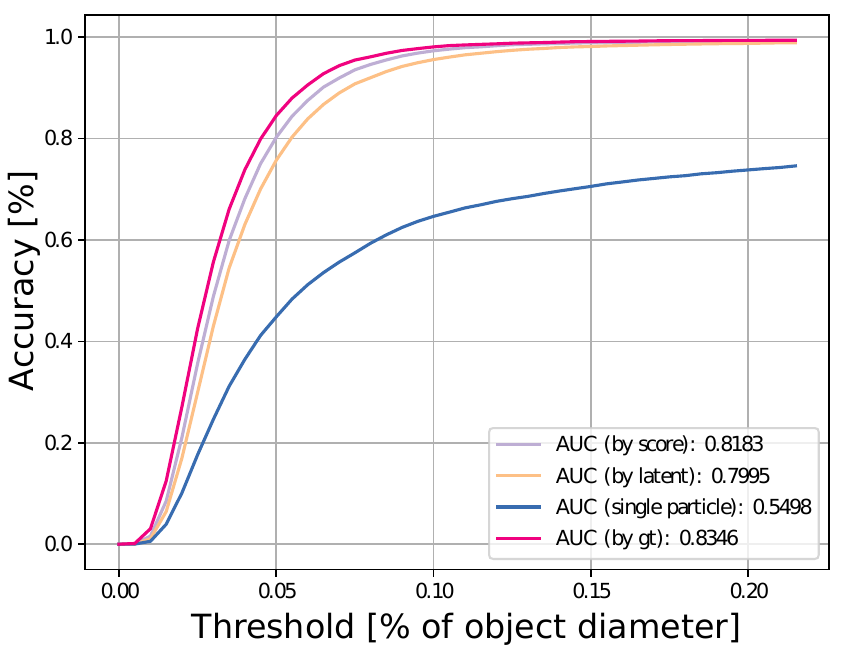}%
	\caption{Accuracy Curve with \AUC{} values for all selection methods for a model trained and tested on all objects. Single particle sampling, i.e. eliminating the need for any selection strategy yields the lower bound and selection by ground truth information the upper bound. %
    }
	\label{fig:selection:onevsmultiobj}
\end{figure}

\begin{figure}	
	\centering
	\includegraphics[width=0.9\linewidth]{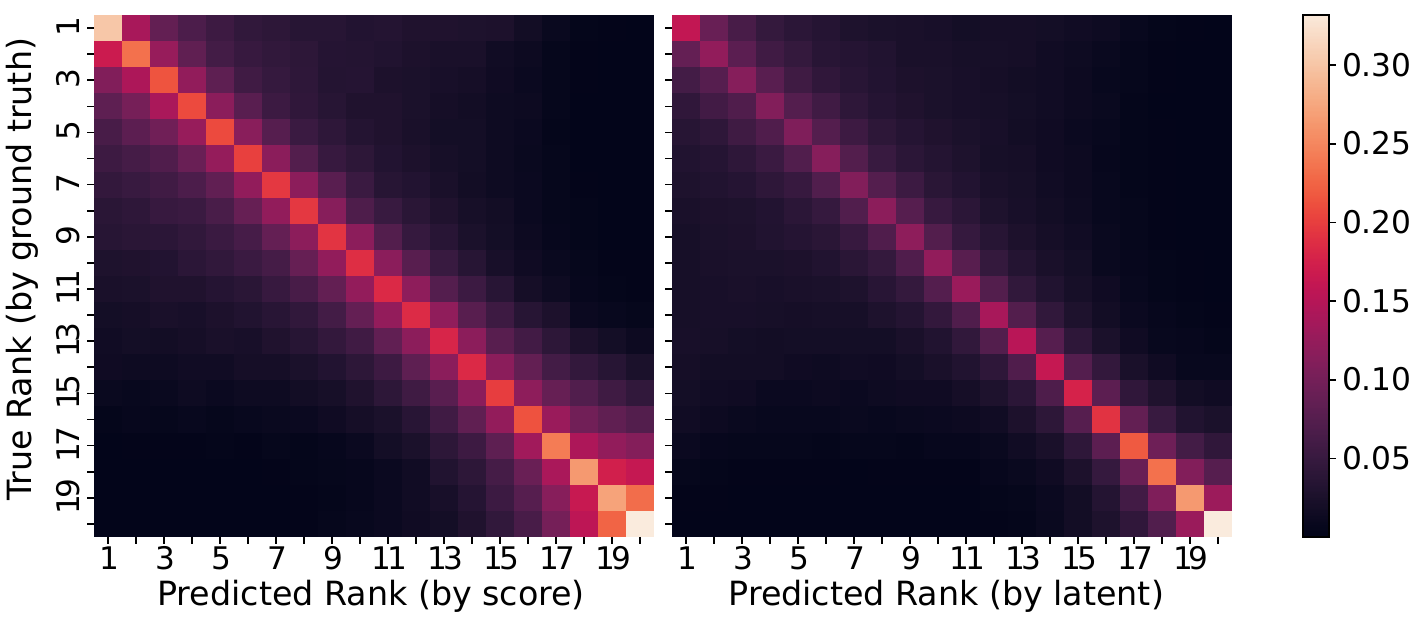}
	\caption{Heatmap showing the correlation between the true rank and the predicted rank of the 20 sampled particles. On the left, the predicted rank is based on the selection by score and on the right side on selection by latent.
 }
	\label{fig:selection_heatmap}
\end{figure}

\begin{table}
\scalebox{0.8}{
	\footnotesize
	\centering
	
	\setlength\extrarowheight{-3pt}
	\begin{tabular}{|c|cc|cc|}%
		\hline
		& \multicolumn{2}{c|}{\textbf{by score}} & \multicolumn{2}{c|}{\textbf{by latent}}\\
		Rank Correlation& statistic & p-value & statistic & p-value \\
		\hline
		Spearman & $0.78 \pm 0.21$ & $0.025 \pm 0.10$ & $0.73 \pm 0.28$ & $0.055 \pm 0.16$ \\
		Kendall & $0.64 \pm 0.21$ & $0.031 \pm 0.12$ & $0.61 \pm 0.28$ & $0.063 \pm 0.18$  \\
		\hline
		
	\end{tabular}
 }
	\caption{
     Comparing the selection strategies for ranking the sampled particles. Their correlation with ground truth rankings is measured using Spearman’s and Kendall’s correlations \& averaged over the test set.
    }
	\label{tab:selection_corr}
\end{table}

To condense the sampled particles, i.e., pose hypotheses, to a single pose prediction, we introduced two selection methods in \cref{sec:twostrategies}: selection by latent and selection by score.
To better understand the performance of our proposed selection strategies, herein, we consider two additional baselines: 1) selection by ground truth (gt), and 2) single particle.
While the selection by ground truth selects the pose hypothesis (particle) with the lowest \gls{ADD}, the single particle baseline naturally eliminates the need for any particle selection strategy since it corresponds to inferring only a single pose using our proposed model.

The results in \cref{fig:selection:onevsmultiobj} show that, as by definition, the selection by ground truth forms the upper performance bound.
In contrast, the single particle baseline forms the lower bound and performs significantly worse compared to all the other particle selection strategies, including the proposed selection by score and latent strategies.
This result underlines the importance of generating multiple pose hypotheses using our diffusion model and subsequently applying particle selection strategies.
One explanation for this observation is that the pose estimation problem is inherently multimodal, and that inferring only a single pose is not sufficient for recovering the mode providing the best pose estimate.
Both our proposed selection strategies lead to comparable results, although the selection by score slightly outperforms the selection by latent with an accuracy (\ADDACC{}) of 97.4\% vs. 95.6\% and a \AUC{} of 81.8 vs. 80.0.
Importantly, they are both much closer to the upper bound than to the lower bound and lead to accurate pose predictions.

For a deeper understanding of the selection strategies, we can interpret them as classifiers, ranking each of the 20 inferred pose hypotheses.
We can compare this ranking with ranking the particles based on the ground truth information.
This way, we can build a confusion matrix shown in \cref{fig:selection_heatmap}.
Ranking by means of selection by score as well as selection by latent are correlated with the true rank as the heat map clearly displays a band alongside the diagonal.
To quantify this observation we compute the Spearman and Kendall correlation between the predicted ranks and the true ranks for each sample of the test set and average them.
As displayed in \cref{tab:selection_corr}, selection by score has a more significant correlation with the true rank compared to the selection by latent (lower p-values), but overall, both show a positive correlation.
As we are mainly interested in the quality of the selected particle with rank 1, i.e. the particle that our model selects as the pose, we examine how often the selected pose is among the top n ranked particles by ground truth.%
In $\approx$57\% of the cases, the predicted pose by score is among the top 3 particles (compared to $\approx$48\% for selection by latent).
When considering the top 5 particles the percentages rise to $\approx$72\% and $\approx$60\% for selection by score \& latent, respectively.
This leads to the conclusion that our particle selection strategies are effective in reliably selecting a pose amongst the best pose candidates and that inferring multiple pose candidates is crucial for obtaining good performance.

\subsection{Partial vs. Full Object Rendering}
\label{sec:partialvsfull}

\begin{figure}
	\centering
	\begin{subfigure}[b]{.5\linewidth}
		\centering
		\includegraphics[width=.9\linewidth]{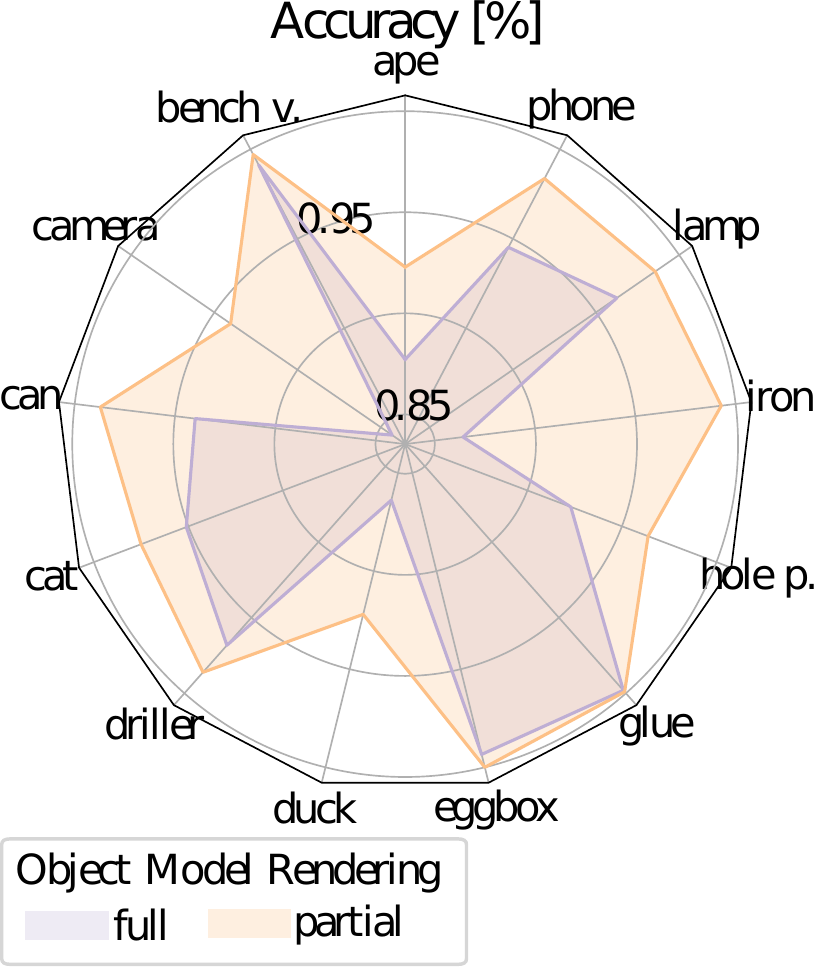}
		\caption{}
		\label{fig:observability_ACC_byscore}
	\end{subfigure}%
	\begin{subfigure}[b]{.5\linewidth}
		\centering
		\includegraphics[width=.9\linewidth]{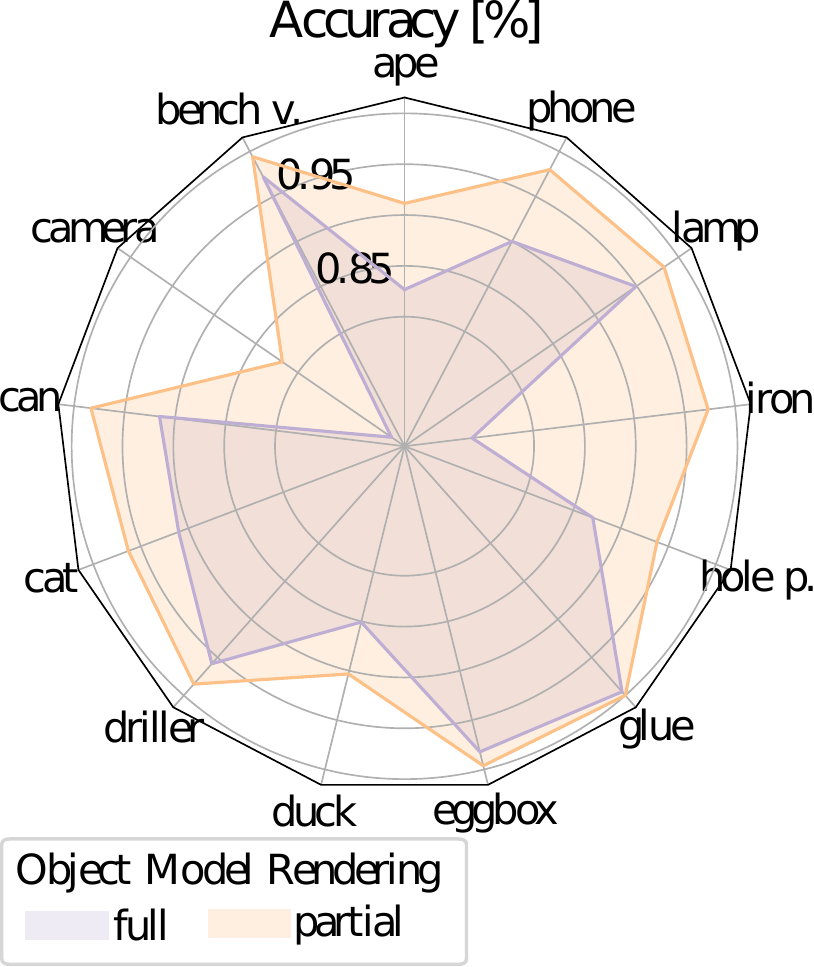}
		\caption{}
		\label{fig:observability_ACC_bylatent}
	\end{subfigure}
	\caption{Comparison of using the partially observable object model and using the full observable object mode.
	(a) is based on selection by score (\cref{sec:selectionbyscore}) whereas (b) is based on selection by latent (\cref{sec:selectionbylatent}). Both plots show the accuracy for each object.
	}
 	\label{fig:observability}
\end{figure}
One design choice of our approach is to only feed the front facing points into the pointcloud encoder to account for partial observability.
This is achieved by the ``partial rendering'' module  (\cref{subsec:partialrendering}).
An alternative would be to use the full transformed object model instead, i.e., the ``full object rendering``.
This experiment, therefore, compares two models.
Our standard model that leverages the partial object rendering, and a baseline model that has been trained using the full object rendering. 

As shown in \cref{fig:observability}, if the final pose prediction is selected by score, we find that considering the partial point cloud is only slightly better in terms of accuracy compared to considering the full object point cloud (97.4\% vs. 93.3\%, respectively).
When examining the accuracy for the distinct objects (\cref{fig:observability_ACC_byscore}), we notice that the partial setting consistently outperforms the full one.
While the performance difference is small for some objects (e.g., bench, glue, eggbox), it is significant for the other objects.
These results are supported by \AUC{} values of 81.8 (partial) vs. 77.5 (full).
Considering selection by latent, we notice a much larger difference between the full and partial observable setting. When using the partially rendered object model, the accuracy is 6.7 percentage points higher (95.6\% vs. 88.9\%) when evaluated across all objects. The accuracy for the distinct objects is consistent with this observation (\cref{fig:observability_ACC_bylatent}). This is supported by the \AUC{} values of 80.0 (partial) vs. 72.9 (full). For additional plots, please see Appendix \cref{app:add_info_partial_rendering}.

While the overall performance of selection by score is better than the selection by latent, it is interesting that there is only little performance difference in the former case compared to a much larger difference in the latter case. As elaborated in \cref{subsec:partialrendering} a full object model leads to a mismatch in the representation of the transformed object and the scene. If selection is performed by latent, this mismatch leads to differences in the latent, although the predicted pose might be good. The partial object model mitigates this problem by filtering those points that are not visible from the current viewpoint. Subsequently, the object latent and the scene latent have a higher similarity for good pose predictions.

\subsection{SE(3)- vs. SO(3)-Equivariant Latent}
\label{sec:se3vsso3}
\begin{figure}
	\centering
	\begin{subfigure}[b]{.5\linewidth}
		\centering
		\includegraphics[width=.9\linewidth]{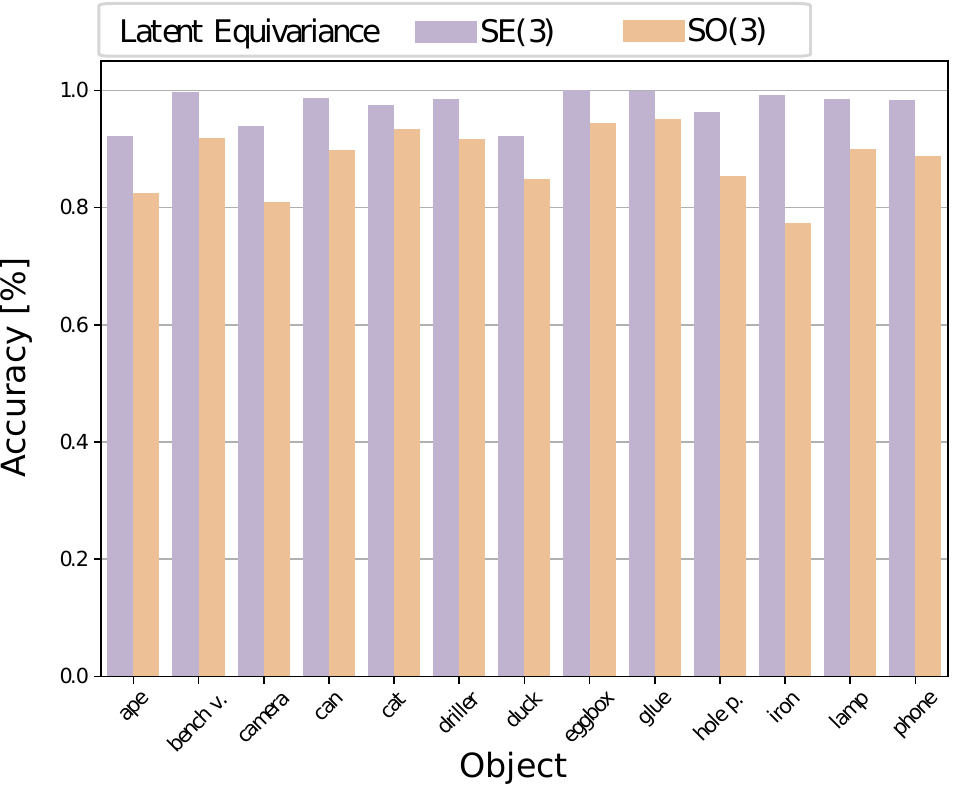}
		\caption{}
		\label{fig:equivariance_byscore}
	\end{subfigure}%
	\begin{subfigure}[b]{.5\linewidth}
		\centering
		\includegraphics[width=.9\linewidth]{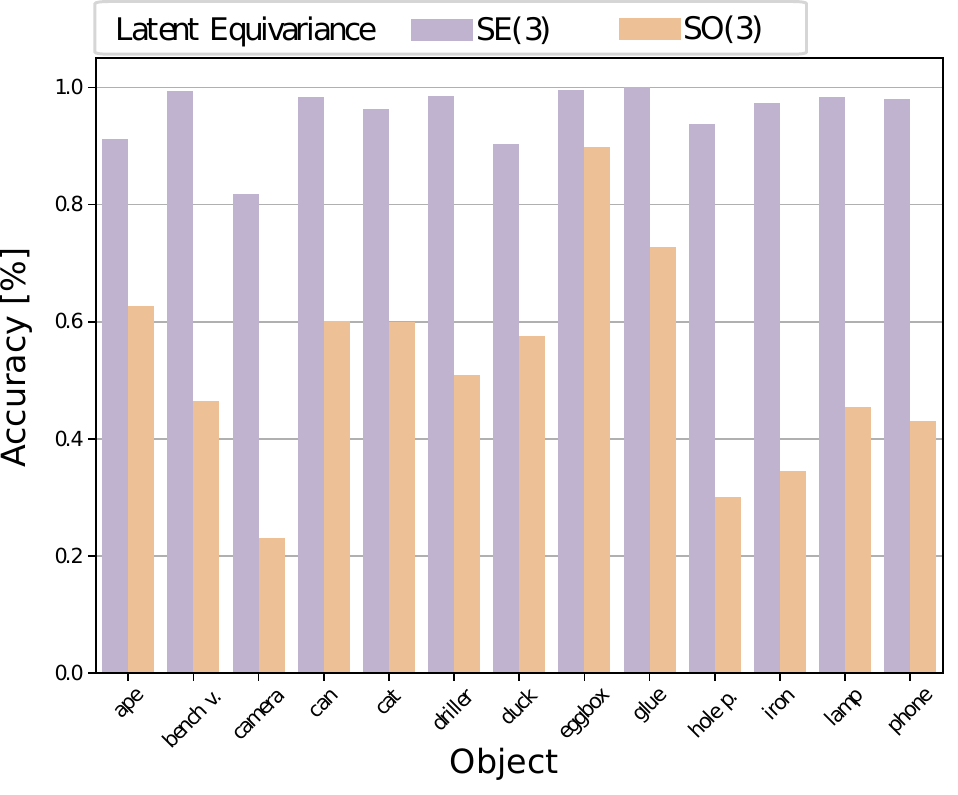}
		\caption{}
		\label{fig:equivariance_bylatent}
	\end{subfigure}
	\caption{Comparison of using a \SE{3}-equivariant latent with its \SO{3}-equivariant alternative. (a) is based on the particle selection by score and (b) based on the selection by latent.}
	\label{fig:equivariance}
\end{figure}
By aligning the cropped scene point cloud through centering before it enters the encoder and then reversing this operation in the latent space, we obtain a \SE{3}-equivariant latent representation (\cref{subsec:pointcloudencoder}).
Omitting the centering and its reversal yields an only \SO{3}-equivariant latent.
This experiment compares the two alternatives by considering an \SO{3}-equivariant baseline model, which has also been trained from scratch.

As shown in \cref{fig:equivariance}, deploying an \SE{3}-equivariant latent improves the performance for both selection strategies. 
For selection by score the accuracy drops by 9.2 percentage points when using the \SO{3}-equivariant latent (mean accuracy of 97.4\% vs. 88.2\%).
In the case of selection by latent, the performance difference is more substantial. The \SO{3}-equivariant latent attains an accuracy of 52.0\% compared to 95.6\% for its \SE{3}-equivariant counterpart.
This discrepancy can be explained by the fact that the selection by latent strategy relies on the assumption that the latent representations of the object and scene are comparable.
However, if the point cloud encoder results in only a \SO{3}-equivariant latent space, changes in the \SE{3} pose might result in more abrupt changes in the latent and, e.g., two poses with equal rotation and only the slightest variation in translation might lead to completely different latents.
We therefore conclude that an \SE{3}-equivariant latent space is an important design choice since it significantly improves the performance regardless of the pose selection strategy.

\subsection{Model Performance}
\label{sec:modelperformance}

\begin{table}
	
	\footnotesize
	\centering
	\scalebox{0.75}{
	\setlength\extrarowheight{-3pt}
	\begin{tabular}{|m{4em}|m{4em}m{4em}m{4em}m{4em}m{4em}|m{4em}|}
		
		\hline
		Approach & PVNet \cite{PVNet} & PoseCNN + DeepIm \cite{PoseCNN, deepIm} & DenseFusion \cite{DenseFusion} &HybridPose \cite{hybridpose} & CloudAAE + ICP \cite{gao2021cloudaae} & \textbf{Ours} \\
		\hline
        Modality & RGB & RGB & RGB-D & RGB & D & D\\ 
        \hline
		ape & 43.6 & \thirdrank{77.0} & \secondrank{92.3}& 63.1 & \firstrank{92.5} &  \secondrank{92.3}\\
		bench v. & \firstrank{99.9} & \thirdrank{97.5} & 93.2 & \firstrank{99.9} & 91.8 & \secondrank{99.7} \\
		camera & 86.9& \thirdrank{93.5} & \firstrank{94.4} & 90.4 & 88.9 & \secondrank{94.0}\\
		can & 95.5 & \thirdrank{96.5} & 93.1 & \secondrank{98.5} & 96.4 & \firstrank{98.7} \\
		cat & 79.3& 82.1& \secondrank{96.5} & \thirdrank{89.4} & \firstrank{97.5} & \firstrank{97.5} \\
		driller & 96.4 & 95.0& 87.0& \thirdrank{98.5} & \firstrank{99.0} & \secondrank{98.6} \\
		duck & 52.6& 77.7 & \secondrank{92.3}& 65.0& 
  \firstrank{92.7} & \thirdrank{92.2} \\
		eggbox* & \thirdrank{99.2} & 97.1 & \secondrank{99.8} & \firstrank{100.0}& \secondrank{99.8} & \firstrank{100.0} \\
		glue* & 95.7 & \thirdrank{99.4} & \firstrank{100.0}& 98.8 & 99.0 & \secondrank{99.9} \\
		hole p. & 81.9& 52.8& \thirdrank{92.1} & 89.7 & \secondrank{93.7} & \firstrank{96.4}\\
		iron & \thirdrank{98.9}& 98.3& 97.0 & \firstrank{100.0}& 95.9 & \secondrank{99.2}\\
		lamp & \secondrank{99.3} & 97.5 & 95.3& \firstrank{99.5} & 96.6 & \thirdrank{98.6} \\
		phone & 92.4 & 87.7 & \thirdrank{92.8} & 84.9& \secondrank{97.4} & \firstrank{98.4}\\
		\hline
		MEAN & 86.3 & 88.6 & \thirdrank{94.3} & 91.3 & \secondrank{95.5} & \firstrank{97.4} \\
		\hline
	\end{tabular}
 }
	\caption{Evaluation and comparison of our approach (using selection by score) with other state-of-the-art approaches for 6D pose estimation on the Linemod dataset. The reported metric is the \ADDACC{} (\cref{sec:metrics}), i.e., accuracy, and colors indicate the three best ranked methods - \firstrank{blue} indicates the best, \secondrank{orange} the second best and \thirdrank{violet} the third best. 
    * denotes symmetric objects. 
    }
	\label{tab:expmodelperformance}
\end{table}

We now compare our model's performance (selection by score \& partial rendering) against other approaches on the Linemod dataset.
As shown in \cref{tab:expmodelperformance}, our proposed methodology demonstrates a competitive accuracy, successfully predicting the correct pose in 97.4\% of cases. Notably, we achieve an accuracy exceeding 90\% for all objects. Particularly noteworthy is our model's performance on symmetric objects, where we achieve nearly 100\% accuracy. This underlines our model's proficiency in learning multi-modalities, especially when faced with ambiguous poses.

Putting the results into the context of alternative approaches, it's important to note that our evaluation, as well as the evaluation for CloudAAE \cite{gao2021cloudaae} was conducted with the advantage of having access to the ground truth segmentation masks for the objects in the test set.
Comapred to CloudAAE, our approach yields an increased mean performance of almost 2.0 percentage points.
However, this assumption is not made by the other methods listed in \cref{tab:expmodelperformance}.
Keeping this advantage in mind, we surpass the performance of DenseFusion \cite{DenseFusion} by 3.1 percentage points and the other baselines even more significantly.
If we run our pose estimator with the object masks obtained from running the same Mask-RCNN as in \cite{cai2022ove6d}, the mean performance drops to 82.9 percent.
We want to point out that this result is still comparable to the most related point cloud-based baseline OVE6D (without running any additional ICP) \cite{cai2022ove6d}, achieving on average 86.1 percent accuracy.
Especially when taking into account that this baseline model was trained on more than 19,000 synthetic objects, and that our own model has only been trained on the perfect segmentation masks from the dataset.

\subsection{Inference Hyper-Parameters and Runtime Analysis}
\label{sec:inferencehyper}

The inference process as detailed in \cref{sec:implementationDetails} encompasses various hyper-parameter.
This section particularily investigates the influence of the rendering interval $k$ on the pose estimation quality and the inference time.
As a reminder, rendering the object model only in every $k$th iteration is only possible due to the \SE{3} equivariant latent, i.e., in all iterations without rendering, we solely transform the object latent which is significantly more efficient compared to re-rendering and re-encoding the updated object point cloud.
For ablations w.r.t. other parameters, please see Appendix \cref{app:additional_inference_hyperparams}.
Experiments are again conducted on an NVIDIA A100 GPU with 40GB.

\begin{figure}
	\centering
	\includegraphics[width=.7\linewidth]{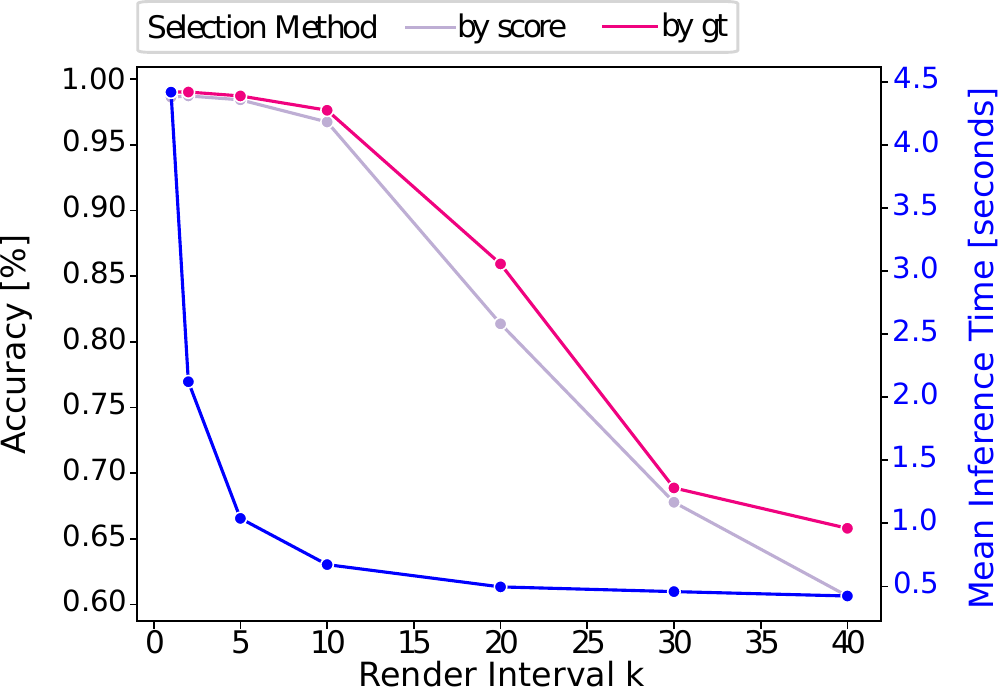}
	\caption{
 Relation between the rendering interval $k$, accuracy, and inference runtime. The interval $k$ controls how often the point cloud is rendered and encoded during inference; in other iterations, only the latent vector $\boldsymbol{z}_o$ is transformed. Results are based on 100 iterations, 20 sampled particles, and the driller object.
	\label{fig:render_intACC}
 }
\end{figure}

This first experiment solely considers the driller object and investigates the effect of varying the render interval $k$ between 1 and 40.
The results are displayed in \cref{fig:render_intACC}. Rendering the object point cloud every iteration leads to an \ADDACC{} of 98.6\% with an inference time of 4.41 seconds per pose prediction. Increasing the render interval to 5 and 10 leads to a similar accuracy while reducing the inference time to 1.04 ($-$76\%) and 0.67 ($-$85\%) seconds.
Higher render intervals heavily reduce the accuracy, and the effect of faster runtimes also vanishes.
For instance, increasing the interval from 10 to 20 reduced the accuracy from 96.7\% to 81.4\% but only decreased the runtime from 0.67 to 0.50 ($-$25\%) seconds. 
Recognizing a favorable trade-off between inference time and accuracy at a rendering interval of $k=10$, we compare this runtime-efficient setting with the default interval of $k=1$ across all objects.
Across all objects, the higher rendering interval only results in a performance decrease of $-$3\%, while significantly improving the inference time by $85$\%.
To further counteract the performance decrease with higher intervals, this effect might need to be reflected in the training procedure.

\section{Conclusion}
\label{chap:conclusion}

This work introduced a novel approach for 6D pose estimation from single-perspective depth images.
To account for the fact that partial observability and symmetric objects yield settings in which multiple pose hypotheses might fit the observation well, this work proposed to train a diffusion-based generative model for pose estimation.
In terms of model architecture, we incorporated recent advancements in point cloud processing and utilized a point cloud encoder that harnesses Vector Neurons to generate a \SE{3}-equivariant latent space.
During inference, the trained generative model allows for inferring multiple pose hypotheses.
To decide upon a final pose estimate from the multiple hypotheses, we introduced two novel pose selection strategies, one inspired by the score-matching objective and another one exploiting the equivariant latent space.
Importantly, both strategies obviate the need for additional model training or other computationally intensive operations.
Our thorough experimental results on the Linemod dataset demonstrated that sampling multiple pose hypotheses and selecting one of them is crucial and significantly outperforms solely inferring a single pose using the trained generative model.
Moreover, the experiments underlined the importance of leveraging the \SE{3} equivariant latent space.%
The equivariant latent space also allowed us to develop a computationally efficient inference strategy that avoids updating and re-encoding the object's point cloud in every inference iteration.
This strategy resulted in a significant acceleration in inference time with a small decline in accuracy and enhances the practicality of our approach in real-world applications. 
In the future, it would be interesting to extend our approach from object pose estimation to object pose tracking.

\section{Acknowledgments}
This work has received funding from the EU’s Horizon Europe project ARISE (Grant no.: 101135959), and the AICO grant by the Nexplore/Hochtief Collaboration with TU Darmstadt.
The authors also gratefully acknowledge the computing time provided to them on the high-performance computer Lichtenberg II at TU Darmstadt, funded by the German Federal Ministry of Education and Research (BMBF) and the State of Hesse.

{
    \small
    \bibliographystyle{ieeenat_fullname}
    \bibliography{main}
}
\clearpage
\maketitlesupplementary

\section{Additional Information on the Partial Rendering Module}
\label{app:add_info_partial_rendering}

\begin{figure}[t]
	\centering
	\includegraphics[width=0.4\linewidth]{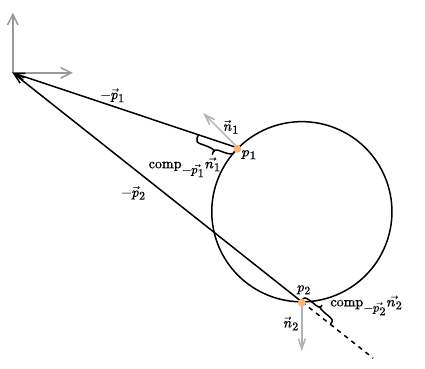}
	\caption{Geometrical visualization of the process to determines whether a point is front facing or not. For simplicity we examine a circle object in 2D. The coordinate system's origin is the viewpoint of the camera and we set the margin $\epsilon=0$. Point $p_1$ is front facing, because the component of it's normal vector in the directional vector from $p_1$ to the origin is positive. At the same time $p_2$ faces backwards because the component of the $\vec{n_2}$ in $-\vec{p}_2$ is negative.}
	\label{fig:methodsPartialRendering}
\end{figure}

We provide \cref{fig:methodsPartialRendering} visualizing the core idea behind the partial rendering module that has been presented in \cref{sec:model_architecture}.

\section{Additional Information on the Point Cloud Encoder}
\label{app:add_information_point_cloud_enc}

\begin{figure}[t]	
	\centering
	\includegraphics[width=0.925\linewidth]{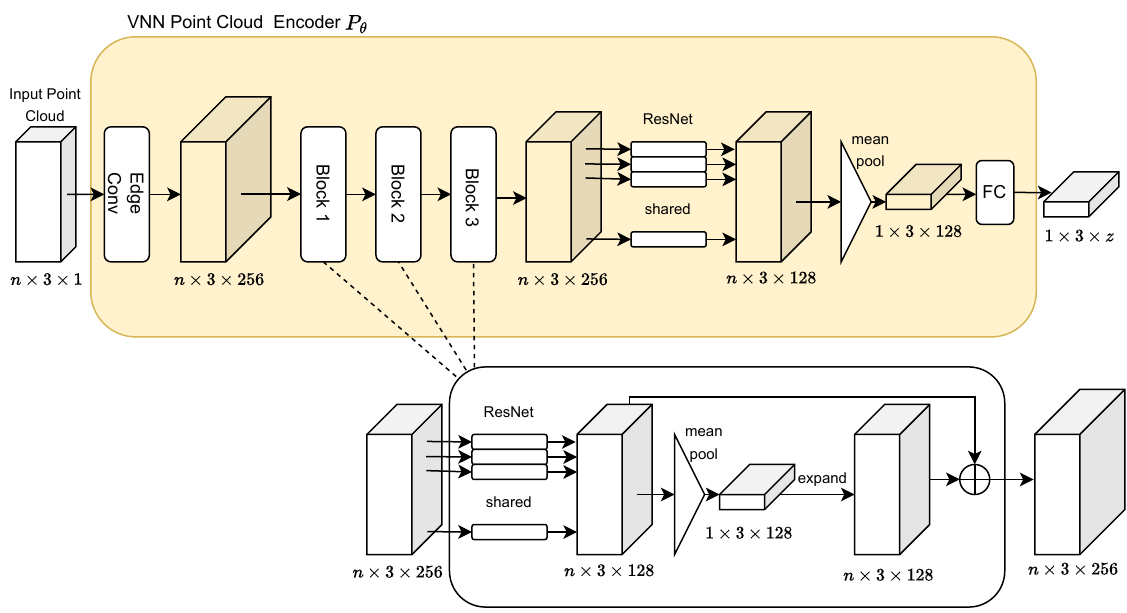}
	\caption{Point Cloud encoder implemented through vector neurons leveraging edge convolution and resnet blocks.
 }
	\label{fig:methodsVNNEncoder}
\end{figure}

The point cloud encoder leverages the \SO{3}-equivariant \glspl{VN} proposed in \cite{VectorNeurons} and defines $P_{\theta} \colon \mathbb{R}^{N \times 3} \to\mathbb{R}^{D \times 3}$ with $D$ being the latent dimension (\cref{fig:methodsVNNEncoder}).
We use the encoder architecture proposed for point cloud reconstruction with minor adjustments \cite{VectorNeurons}.
A similar architecture was also used in \cite{DiffusionFields}.
Inspired from DGCNNs an edge convolution layer \cite{DGCNN} is used to compute features from an input point cloud $\boldsymbol{x} \in \mathbb{R}^{N \times 3}$.
In doing so, for each point the 20 nearest neighbors are computed forming a $20 \times 3$ matrix that is processed by a non-linearity to form a edge feature of shape $3 \times 256 \times 20$.
Afterwards mean pooling is applied on the edge features to create a $3 \times 256$ descriptor of the point.
For more details refer to \cite{DGCNN}.
The point descriptors are subsequently processed by multiple identical processing blocks.
The blocks' architecture is inspired by PointNet \cite{PointNet} and has a shared Residual Network at its core.
The output of the third block is processed by another Residual Network, aggregated and fit to the expected embedding dimension $D$.

\section{Inference Process - Pseudo Code}
\label{app:inference_pseudocode}

This section provides the pseudo code for our inference procedure, i.e., cf. \cref{alg:ourannealed}.

\begin{algorithm}
\scriptsize	
	\caption{Langevin Dynamics Inference.}
	\label{alg:ourannealed}
	\begin{algorithmic}[1]
		\Require 
		\Statex $\theta$: parameters for $P_{\theta}$, $F_{\theta}$, $D_{\theta}$, $T_{\theta}$
		\Statex $(t_i)_{i=0}^{L-1}$: sequence of time steps
		\Statex $ \boldsymbol{x}_s, \boldsymbol{x}_o, \boldsymbol{n}_o$: scene point cloud, object point cloud, object normal vectors
		\Statex $\sigma$: noise scale hyper-parameter
		\Statex $k$: rendering interval
		\State Initialize $\boldsymbol{H}_{0}=\Expmap(\boldsymbol{\epsilon}) \quad \boldsymbol{\epsilon} \sim \mathcal{N}(0,\textrm{var}(\sigma, 1)\boldsymbol{I}) $
		\State $\boldsymbol{z}_{s} = P_{\theta}(\boldsymbol{x}_s)$ \Comment{encode scene point cloud}
		\For{$l \leftarrow 0$ to $L-1$} 
		
		\If{$l \mod k$} 
			\State $\boldsymbol{z}_{o,l}=P_{\theta}(\mathrm{partialrender}(\boldsymbol{x}_o, \boldsymbol{n}_o, \boldsymbol{H}_l))$ \Comment{encode object point cloud}
		\Else
			\State $\boldsymbol{z}_{o,l}=\boldsymbol{H}_l \boldsymbol{z}_{o,l-1}$ \Comment{transform object latent}
		\EndIf
		
		\State $\boldsymbol{\psi}_l = (\mathrm{Flatten}(\boldsymbol{z}_s), \mathrm{Flatten}(\boldsymbol{z}_{o,l}))$ \Comment{flatten and concatenate latents}
		\State $\boldsymbol{s}_l = D_{\theta}(F_{\theta}(\boldsymbol{\psi}_l, \boldsymbol{H}_l, T_{\theta}(t_l)))$ \Comment{compute score}
		\State $\alpha_{l} = 2\left\Vert  \boldsymbol{s}_{l} \right\Vert_2^{-2}$
		\If{$l<(L-5)$} \Comment{Langevin dynamics update with noise}
			\State $\boldsymbol{H}_{l+1}= \Expmap(\alpha_{l} \boldsymbol{s}_l+ \sqrt{2\alpha_{l}}0.01\boldsymbol{\epsilon})\boldsymbol{H}_{l} \quad \boldsymbol{\epsilon} \sim \mathcal{N}(\boldsymbol{0},\boldsymbol{I})$ 
		\Else \Comment{Langevin dynamics update without noise}
			\State $\boldsymbol{H}_{l+1}= \Expmap(\alpha_{l} \boldsymbol{s}_l)\boldsymbol{H}_{l}$ 
		\EndIf
		
		\EndFor 
		\Return $\boldsymbol{H}_{L}$
	\end{algorithmic}
	
\end{algorithm}

\section{Additional Experimental Results}

\subsection{Selection Methods when Training a Single Object Model}

While \cref{sec:selection_methods} compared the two selection strategies when training a single model for object pose estimation with all of the objects, this section provides additional results for the selection strategies when training a model for a single object (i.e., the driller).
In this case, we observe that the \ADDACC{} of selection by latent remains unaffected while selection by score leads to a much lower \ADDACC{} of only 72.7\% compared to 97.4\%.
Therefore, we reason that selection by score seems to profit from multi-object training while selection by latent is not as dependent on the utilized number of objects during training.

\subsection{Additional Plots for Full vs. Partial Object Rendering}

\begin{figure}
	\centering
	\begin{subfigure}[b]{.5\linewidth}
		\centering
		\includegraphics[width=.9\linewidth]{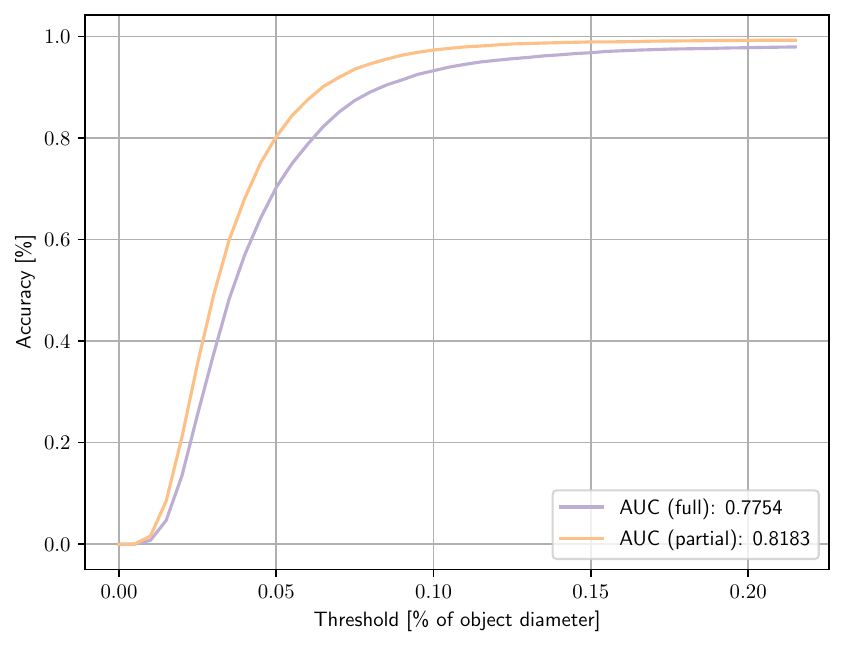}
		\caption{}
		\label{fig:observability_AUC_byscore}
	\end{subfigure}%
	\begin{subfigure}[b]{.5\linewidth}
		\centering
		\includegraphics[width=.9\linewidth]{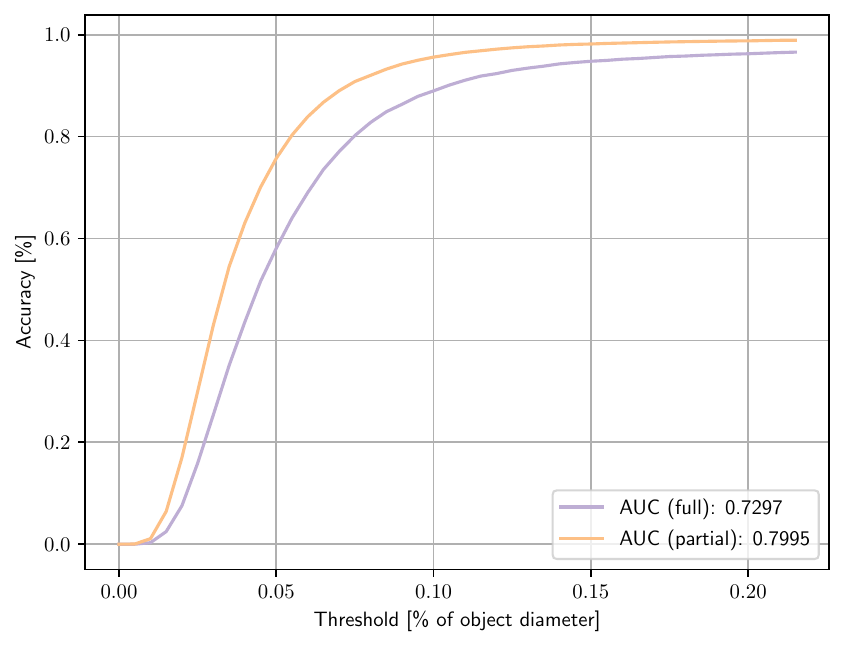}
		\caption{}
		\label{fig:observability_AUC_bylatent}
	\end{subfigure}
	\caption{Comparison of using the partially observable object model and using the full observable object mode.
	(a) is based on selection by score (\cref{sec:selectionbyscore}) whereas (b) is based on selection by latent (\cref{sec:selectionbylatent}). Both plots show the accuracy curve for different thresholds together with the \AUC{}  metric.
	}
 	\label{fig:app_auc_observability}
\end{figure}

This section provides additional plots regarding the full vs partial rendering experiment that has been presented in \cref{sec:partialvsfull}.
In particular, \cref{fig:app_auc_observability} depicts the plots for obtaining the AUC values.

\subsection{Additional Results on Inference Hyper-Parameters and Runtime Analysis}
\label{app:additional_inference_hyperparams}

The inference process as detailed in \cref{sec:implementationDetails} encompasses various hyper-parameters, like the number of iterations, the number of particles to draw or the rendering interval $k$.
While in the main paper, we focus on the rendering interval, herein we provide additional results w.r.t. the number of iterations and the number of particles to draw.
The following experiment investigates the effect of different settings for those parameters on the pose estimation quality, and the inference time.
All following inferences are conducted on the same hardware, a NVIDIA A100 GPU with 40GB.
For the scope of the following experiments, the evaluation is performed on only the driller object (Obj 8).

\begin{figure}
	\centering
	\includegraphics[width=.85\linewidth]{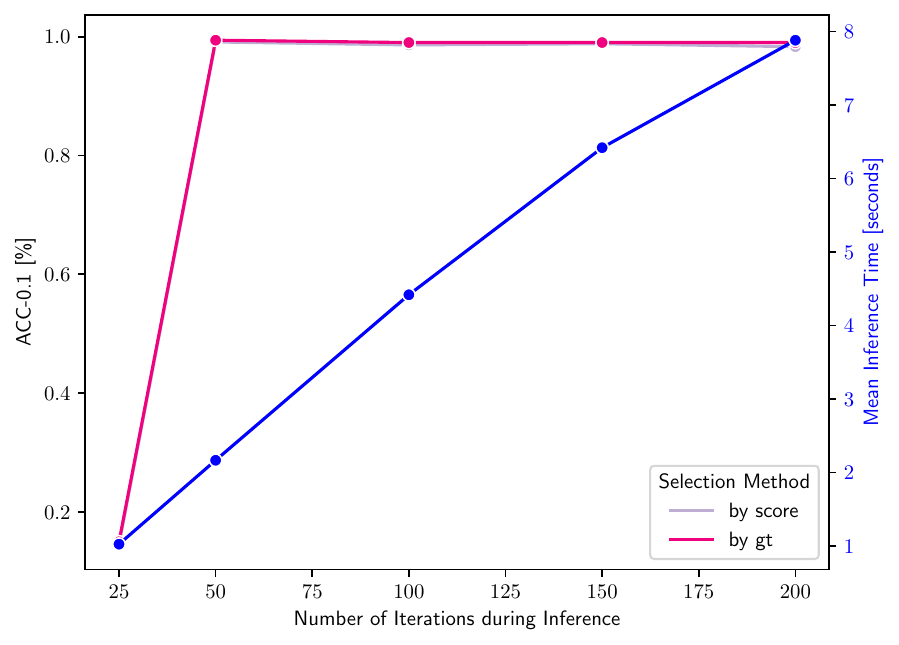}
	\caption{Relation between number of iterations deployed in the inference process and the accuracy on the driller object as well as the connection to the inference time. The inference time is averaged over all samples of the dataset. Results are based on 20 sampled particles with a render interval of $k=1$.}
	\label{fig:n_iterationsACC}
\end{figure}

\textbf{Number of Inference Steps.} The inference process needs to be deployed with a sequence of $L$ time steps starting at 1 all the way down to a value close to 0. Yang et al. \cite{YangImproved} recommend to choose $L$ as large as allowed by the labeling budget. We fix the number of sampled particles to 20 and the rendering interval to $k=1$.
With $L=25$  the \ADDACC{} is as low as $\approx$15\%. However, with 50 iterations the accuracy is already at $\approx$99\% and is not improved with a further increase in the number of iterations. For 100, 150, and 200 iterations the accuracy remains at the same level. This relation holds for both sampling strategies. As visualized in \cref{fig:n_iterationsACC}, we observe a linear relation between the number of iterations and the required inference time per sample. With 25 iterations, the inference time amounts to 1.03 seconds and increases to 2.17 seconds for 50 iterations. 

\begin{figure}
	\centering
	\includegraphics[width=.85\linewidth]{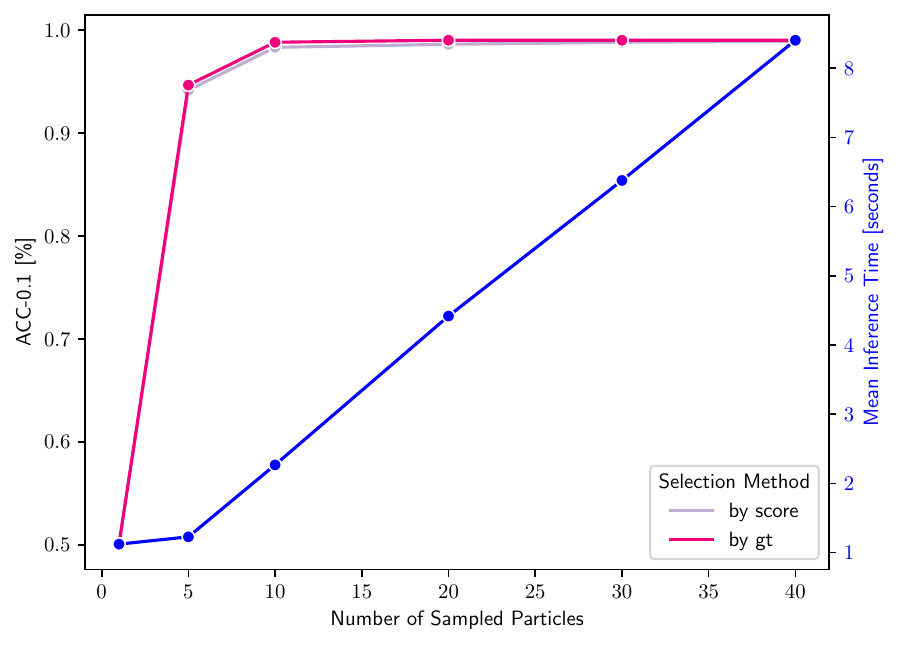}
	\caption{Relation between number of sampled particles and pose prediction accuracy including the corresponding inference time. Results are based on 100 iterations and a rendering interval of $k=1$ during inference.}
	\label{fig:n_samplesACC}
\end{figure}
\textbf{Number of Particles.} All experiments in the paper are based on 20 drawn pose hypotheses, from which we used the two proposed pose selection strategies to make the final pose prediction. 
In this experiment the number of drawn particles is varied while the number of iterations is set to 100 and the rendering interval to $k=1$. If we only draw one particle, our approach achieves an accuracy of 50\%. Increasing the number of sampled particles to 5, 10, and 20 increases the \ADDACC{} further to respectively 94\%, 98\%, 99\% (\cref{fig:n_samplesACC}). A further rise is not accompanied by an increase in accuracy but rather leads to stagnation. 
From 1 to 5 drawn particles, the inference time makes a minor jump from 1.12 to 1.22 ($+$9\%) seconds. Afterwards, a duplication in sampled pose hypothesis leads to an increase of $\approx$90\% in inference time. The number of drawn samples is the main driver that determines the required GPU size.

\textbf{Rendering interval of $k{=}10$ on all objects.}
\cref{tab:optimalhyperparams} provides a detailed comparison when running our proposed model with a rendering interval of $k{=}10$ vs. rendering in every iteration ($k{=}1$) on all objects.

As shown in \cref{tab:optimalhyperparams}, and as expected, for both symmetric objects, we basically do not observe any drop in accuracy. Due to their symmetry, when rotating, re-rendering, and re-encoding them, the latent embeddings change the least, and therefore, only rendering every kth iteration basically leaves the outcome of our pose estimator unchanged.
Across all objects, the drop in accuracy is small (only 2.8\%); however, for some of the objects, particularly the duck, the accuracy decreases by 8.2 percentage points.

\begin{table}
	
	\footnotesize
	\centering
	
	\setlength\extrarowheight{-3pt}
	\begin{tabular}{|m{4em}|m{3em}m{6em}|m{3em}m{6em}|} 
		\hline
		& \multicolumn{2}{c|}{\textbf{\ADDACC}} & \multicolumn{2}{c|}{\textbf{\AUC{}}}\\
		Object & $k=1$& $k=10$ & $k=1 $ & $k=10$ \\
		\hline
		ape & 92.3 & 86.6 ($-5.7$)& 74.6 & 70.1 ($-4.5$) \\
		bench v. & 99.7 & 98.4  ($-1.3$) & 86.1 & 82.4 ($-3.7$) \\
		camera & 94.0 & 88.6 ($-5.4$) & 73.9 & 69.3 ($-4.6$) \\
		can & 98.7 & 95.7 ($-3.0$) & 83.0 & 79.0 ($-4.0$) \\
		cat & 97.5 & 95.8 ($-1.7$) & 82.2 & 78.8 ($-3.4$) \\
		driller & 98.6 & 96.7 ($-1.9$) & 85.6 & 81.8 ($-3.8$)\\
		duck & 92.2 & 84.0 ($-8.2$) & 72.5 & 65.9 ($-6.6$) \\
		eggbox* & 100 & 100 ($-0.0$) & 86.2 & 84.6 ($-1.6$) \\
		glue* & 99.9 & 99.8 ($-0.1$) & 88.6 & 86.6 ($-2.0$) \\
		hole p. & 96.4 & 92.4 ($-4.0$) & 77.8 & 73.6 ($-4.2$) \\
		iron & 99.3 & 97.8 ($-1.5$) & 84.6 & 81.2 ($-3.4$) \\
		lamp & 98.6 & 97.6 ($-1.0$) & 86.8 & 83.4 ($-3.4$)\\
		phone & 98.4 & 96.1 ($-2.3$) & 82.4 & 78.7 ($-3.7$) \\
		\hline
		MEAN & 97.4 & 94.6 ($-2.8$) & 81.9 & 78.1 ($-3.8$) \\
		\hline
		
	\end{tabular}
	\caption{Comparison in performance of rendering the object every iteration ($k=1$) with periodically rendering every 10$^{\textrm{th}}$ iteration ($k=10$). Both alternatives are based on the same model that was trained on all objects. Value in parenthesis describes the difference in percentage points.}
	\label{tab:optimalhyperparams}
\end{table}

\end{document}